\documentclass[11pt]{article}

\usepackage[preprint]{acl}

\usepackage{times}
\usepackage{latexsym}

\usepackage[T1]{fontenc}

\usepackage[utf8]{inputenc}

\usepackage{microtype}

\usepackage{inconsolata}

\usepackage{graphicx}

\usepackage{pifont}

\title{SYNTHEMPATHY: A Scalable Empathy Corpus Generated Using LLMs Without Any Crowdsourcing}

\author{Run Chen \\
  Columbia University \\
  \texttt{runchen@cs.columbia.edu} \\\And
  Jun Shin \\
  Columbia University \\
  \texttt{js5810@columbia.edu} \\\And
  Julia Hirschberg \\
  Columbia University \\
  \texttt{julia@cs.columbia.edu}
  }

\begin{document}
\maketitle
\begin{abstract}
Previous research has shown that humans are more receptive towards language models that that exhibit empathetic behavior. While empathy is essential for developing helpful dialogue agents, very few large corpora containing empathetic dialogues are available for fine-tune LLMs. The few existing corpora have largely relied on crowdsourcing to simulate empathetic conversations, a process that is expensive, time-consuming, and not scalable to larger datasets. We propose a data generation framework for developing SYNTHEMPATHY, a large corpus containing 105k empathetic responses to real-life situations compiled through LLM generation. A base Mistral 7B model fine-tuned on our SYNTHEMPATHY corpus exhibits an increase in the average empathy score.
\end{abstract}

\section{Introduction}

Incorporating empathy into dialogue systems fosters trust and likability among users \cite{9473685}. High-quality empathy corpora are crucial for training language models in empathy, as these models typically do not focus on empathy during pre-training and must be fine-tuned to develop empathetic capabilities. Despite their importance, high quality large scale empathy corpora are scarce 
due to challenges such as i)
the scarcity of empathetic texts on the internet, in fact many hostile and anti-empathetic; ii) difficulty in accurately identifying empathetic text within internet data, which poses a `chicken or egg' problem: training an effective model to perform this task requires substantial amounts of empathetic data, which is itself scarce.



To create such an empathetic dataset, researchers have either employed expert annotations \cite{chen2024detecting} or crowdsourcing \cite{rashkin-etal-2019-towards} for reliable labeling.
However, crowdsourcing, while valuable, is not a scalable solution for developing large corpora due to its resource intensity in terms of both time and financial investment \cite{webb2022too}
Additionally, the implementation of crowdsourcing presents practical challenges, as workers on popular platforms like Amazon Mechanical Turk (MTurk) often lack domain expertise in the targeted area and 
may struggle to overcome language barriers necessary for complicated tasks, such as responding empathetically to a distressed person.
Recent studies have even raised ethical concerns about the using crowdsouring in academic research settings \cite{moss2023ethical}.
To the best of our knowledge, all existing large empathy corpora has involved at least one step of crowdsourcing, which has limited their size and range of topics covered by these corpora.
\begin{figure}
  \centering  \includegraphics[width=\linewidth] {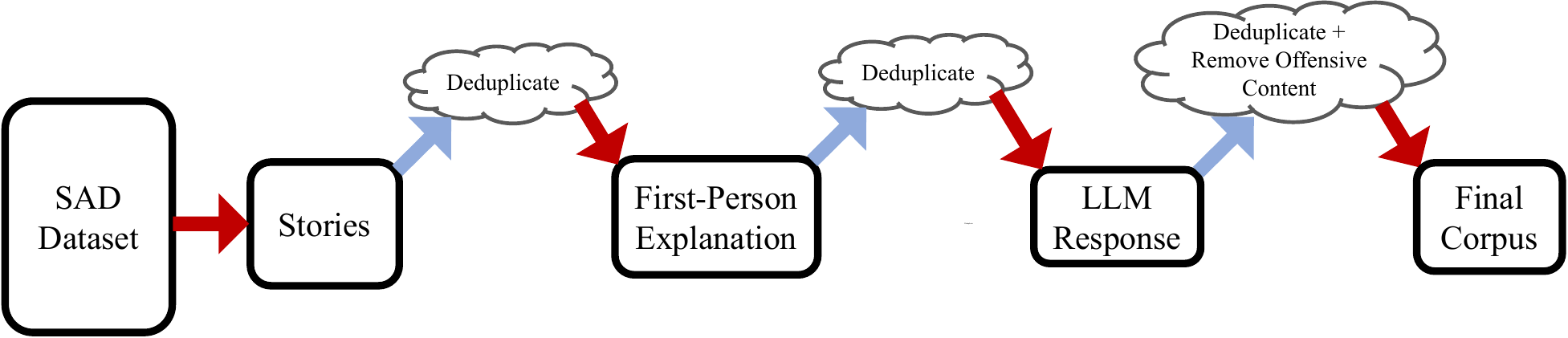}
  \caption{Corpus Construction Pipeline Overview}
  \label{fig:pipeline}
\end{figure}

We propose a novel, self-sufficient framework for constructing an empathy corpus without relying on crowdsourcing. We establish a step-by-step pipeline using Large Language Models (LLMs) to first brainstorm scenarios that warrant empathetic responses and eventually generating such responses through a special prompting method grounded in psychotherapy theories. This approach fully leverages the creative potential of LLMs that can surpass human performance \cite{Girotra2023IdeasAD}. Interestingly, the hallucinatory nature of LLMs is actually helpful here, since it enables the generation of a large repertoire of unique scenarios. A key advantage of this method lies in its scalability, allowing for the creation of a substantially larger corpus without the financial and logistical constraints associated with crowdsourcing. The resulting SYNTHEMPATHY corpus consists of 105,578 empathetic single-turn dialogues generated using this framework. 

Our main contributions are:
1) a novel, step-by-step framework for generating a corpus of empathetic dialogues without any crowdsourcing or web crawling,
2) a large SYNTHEMPATHY corpus containing 105,578 empathetic explanation-response pairs, each grounded in a distinct real-life scenario,
and 3) a Mistral 7B model, fine-tuned on the SYNTHEMPATHY corpus to demonstrate a measurable enhancement in empathetic capabilities\footnote{The code and dataset will be made publicly available upon release of the final version of this paper.}.

\section{Related Work}

Small hand annotated corpora \cite{chen2024detecting} provides useful insights into empathy expression; however, their limited size may not be sufficient for fine-tuning LLMs. We focus on large-scale, textual corpora that are suitable for training and fine-tuning most LMs in use today.


\subsection{Empathy Corpora}
Table~\ref{tab:corpora comparison} provides a comprehensive comparison of key metrics and characteristics of existing empathy corpora as well as SYNTHEMPATHY. 

\begin{table*}
  \centering
  \begin{tabular}{lccccc}
    \hline
     & \textbf{ED} & \textbf{EPITOME} & \textbf{OSED} & \textbf{SoulChatCorpus} & \textbf{SYNTHEMPATHY} \\
    \hline
    Num. Examples & 24k & 10k & 1M & 200k & 105k                          \\
    Utterances per Example & 4.31 & 2.00 & 3.49 & 11.50 & 2.00  \\
    Crowdsourced & \color{green}{\ding{51}} & \color{green}{\ding{51}} & \color{green}{\ding{51}} & \color{green}{\ding{51}} & \color{red}{\ding{55}}      \\
    Topics Evenly Distributed & \color{green}{\ding{51}} & \color{green}{\ding{51}} & \color{red}{\ding{55}} & \color{red}{\ding{55}} & \color{green}{\ding{51}}      \\
    \hline
  \end{tabular}
  \caption{Comparison of key metrics of empathy corpora. Our SYNTHEMPATHY dataset is the first large-scale corpus that excludes crowdsourcing and balances the topic distributions.}
  \label{tab:corpora comparison}
\end{table*}

Earlier efforts in empathetic dataset collection and annotation, such as the EmpatheticDialogues (ED) \cite{rashkin-etal-2019-towards} and EPITOME \cite{sharma-etal-2020-computational} have predominantly relied on crowdsourcing. Specifically, ED is a multi-turn empathy corpus, assembled by engaging 810 Amazon MTurk workers to chat in pairs, each conversation prompted by one of 32 assigned emotion labels. 

The EPITOME empathy corpus \cite{sharma-etal-2020-computational} was created by web crawling from Reddit and the online mental health forum TalkLife, and subsequently annotated through crowdsourcing, which required fewer crowdsourced workers. Eight crowdsourced workers evaluated the post-response pairs by scoring each each based on how well it expressed emotional reaction (ER), interpretation (IP), and exploration (EX).
ER is a crucial indicator of empathy as revealing one's own emotions can foster empathetic rapport with the original poster. IP signals understanding of the poster's struggles, paving the way for deeper empathetic connections. Lastly, EX suggests new perspectives on the seeker’s experience, crucial for conveying empathy-driven interest.
These three empathetic metrics are similar to practices used in psychotherapy \cite{fuller2021conceptualizing, jani2012role, chen2024detecting}.

More recently, researchers have combined crowdsourcing with further extrapolation using LLMs to expand dataset sizes.
\citet{welivita2020finegrained} developed the OpenSubtitles Emotional Dialogue (OSED) by extracting 1M dialogues from movie subtitles. Each dialogue contains utterance-level labels for emotion and empathetic intent, assigned by a BERT-based classifier fine-tuned on a development set of 9k dialogues, which had been manually corrected by Amazon MTurk workers. Due to the high cost of scaling crowdsourcing, only about 0.91\% of the dialogues were manually checked, underscoring the challenge of expanding manual checks in large datasets.
The SoulChatCorpus \cite{chen-etal-2023-soulchat} was built by initially collecting 215,813 question-answer pairs through crowdsourcing, followed by utilizing ChatGPT as a rewriting tool to transform each pair into a multi-turn dialogue. Each dialogue ranges from 8 to 20 turns, resulting in approximately 2M utterances. Both SoulChatCorpus and OSED are limited by their reliance on crowdsourced workers to create an initial high-quality subset of the corpus.

\subsection{Empathy Generation}

Previous efforts to generate empathetic responses from LLMs have involved modifying the underlying model architecture, fine-tuning on empathy corpora, or employing meticulous prompting to improve empathy levels of the outputs.
Adding emotion tags or emotional embeddings
\cite{rashkin-etal-2019-towards, goel2021emotion} improves response generation. 
\citet{lee2022improving} attached a normal distribution random sampler right before the decoder in order to inject more stochasticity into empathetic dialogue agents making its empathetic responses sound personalized. Due to the lack of large empathy corpora, very few studies focus on fine-tuning. \citet{chen-etal-2023-soulchat} fine-tuned ChatGLM-6B on the SoulChatCorpus corpus to determine how much the base model improves. 

Prompt engineering, especially Chain of Thought prompting, is increasingly popular in enhancing LLMs for downstream tasks in zero-shot or few-shot settings \cite{wei2022chain}. LLMs generate more empathetic responses when the prompts incorporate psychotherapy approaches used by professional therapists, that is the Chain of Empathy (CoE) prompting \cite{lee2023chain}. 
 This approach involves step-by-step prompts that not only describe a client’s situation but also include reasoning for why empathy is needed, modeled after various therapeutic styles including Cognitive Behavioral Therapy (CBT), Dialectical Behavior Therapy (DBT), Person-Centered Therapy (PCT), and Reality Therapy (RT). Further details on each therapy approach are available in Appendix A.1. Our pipeline incorporates \citeposs{lee2023chain} CoE prompting as a crucial component.
 
\section{Corpus Construction Framework}
Our framework alternates between generation and deduplication steps in sequence (Figure \ref{fig:pipeline}). The SYNTHEMPATHY corpus is produced by a step-by-step process of story brainstorming, explanation rewriting, and empathetic response. We refer to these three steps as the generation steps. In this process, we run an assortment of LLMs, including Llama 2 13B Chat, Llama 3 8B, and Gemma 7B, to enhance diversity and minimize repetition in the output texts used in the subsequent step. We maintain the corpus quality by implementing routine deduplication steps in between each generation step. Furthermore, the last deduplication step includes a manual keyword search to remove any examples with offensive language. 

\begin{figure}
  \centering
  \includegraphics[width=\linewidth] {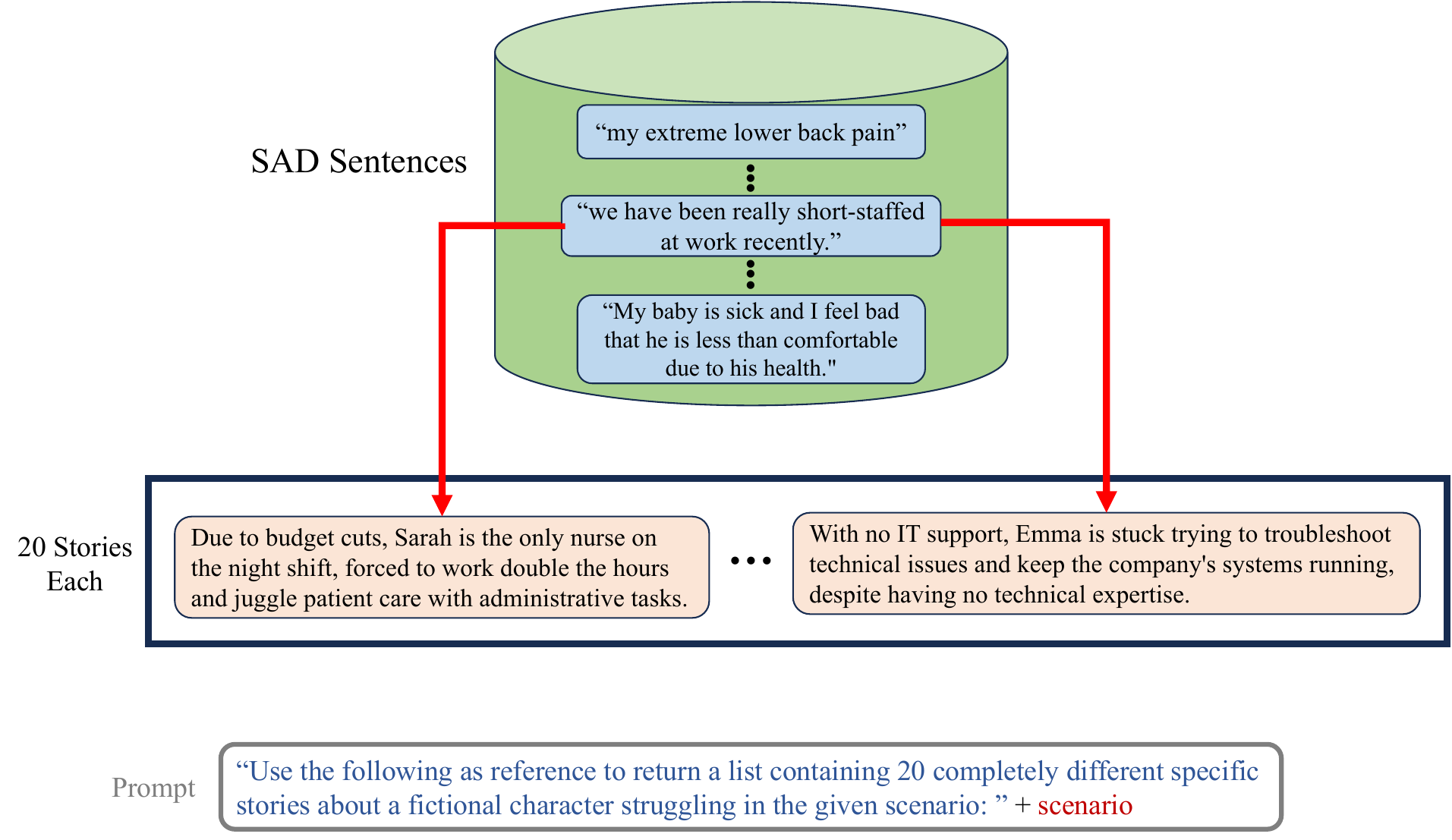}
  \caption{Story Brainstorming Step. Each sentence from the SAD dataset \cite{10.1145/3411763.3451799} is prompted into Llama 2 13B Chat to generate 20 stories.}
  \label{fig:storybrainstorming}
\end{figure}

\subsection{Story Brainstorming}
The first step in our pipeline involves generating stories based on various scenarios. We apply the scenarios from English Stress Annotated Dataset (SAD) \cite{10.1145/3411763.3451799}, which contains stress-inducing scenarios that are annotated with many features including severity ratings.
As shown in Figure \ref{fig:storybrainstorming}, Llama 2 13B Chat uses each of the 6,476 SAD scenarios as inspiration to generate 20 unique stories, and thus creating 129,520 stories in total. The optimal values for the hyperparameters \textit{temperature} and \textit{top\_p} are determined empirically through a simple grid search while setting the temperature high to maximize randomness. Further details on hyperparameter tuning can be found in Appendix \ref{appendix:HyperparameterTuning}.

\subsection{Story Deduplication}
We employ the \textit{ExactSubstr} algorithm \cite{lee-etal-2022-deduplicating}, which uses a suffix array to efficiently identify and remove all substring matches across the input data in mostly $O(\log N)$ operations. We set the number of characters that must match before the algorithm removes it, $\textit{dup\_length\_threshold}$ to 75. The algorithm trims our stories by 12\%, removing 14,863 duplicate stories and leaving us with 114,657 unique stories suitable for rewriting as first-person explanations in the next generation step.

\subsection{Explanation Rewriting}
\begin{figure}
  \centering
  \includegraphics[scale=0.23] {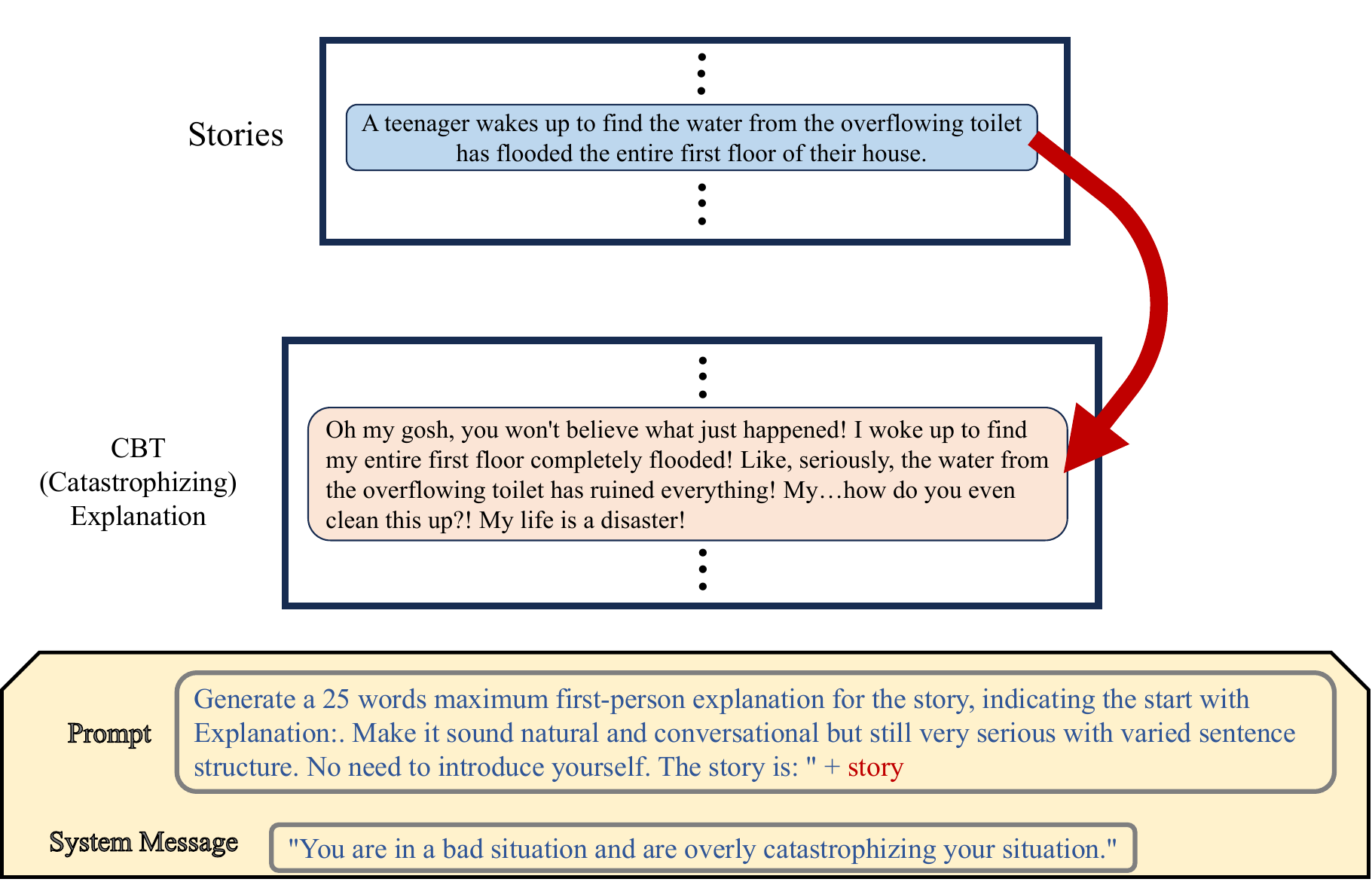}
  \caption{Explanation Rewriting Step. Each story is rewritten into a first-person explanation. This figure illustrates an example story being converted into a Cognitive Behavioral Therapy (CBT) explanation.}
  \label{fig:ExplanationRewriting}
\end{figure}

We use LLMs as text rewriting tools to convert each story into a first-person explanation designed to elicit empathetic responses. Given the four types of Chain of Empathy (CoE) approaches \cite{lee2023chain}, namely CBT, DBT, PCT, and RT, we divide the stories into four equal bins and generate the first-person explanations corresponding to one of the therapy types by varying the system message. Figure \ref{fig:ExplanationRewriting} shows the prompt and system message input to the LLMs.

\subsection{Explanation Deduplication}
As the previous story deduplication, we run the \textit{ExactSubstr} algorithm, which removes 165,365 duplicate characters across the first-person explanations.

\subsection{Empathetic Response}
The final step in our pipeline involves feeding the deduplicated CoE explanations to an LLM to retrieve empathetic responses. We use an even mix of Llama 2 13B Chat, Gemma 7B, Mistral 7B, and Llama 3 8B for a variety of response styles. After collecting all explanation-response pairs, we conduct one last deduplication step using \textit{ExactSubstr} algorithm with $\textit{dup\_length\_threshold}$=100 characters. We then remove any offensive content using keyword searches. Ultimately, we obtain the SYNTHEMPATHY corpus of 105,578 explanation-response pairs. The explanations have a mean length of 95.5 words and a standard deviation of 80.8 words, while the responses have a mean of 153.7 words and a standard deviation of 69.8 words (See Appendix Figure \ref{fig:data-stats}).

\section{Results}

To explore the potential of the SYNTHEMPATHY corpus in improving LLMs, we fine-tune\footnote{adapted from a public Python notebook on unsloth.ai, which provides kernel-level optimizations for LLM fine-tuning} Mistral 7B using this corpus (\textbf{Fine-Tuned}) and compare the responses it produces against those generated by the base Mistral 7B model (\textbf{Base}) in Table~\ref{tab:empathy score}. 

We test both models on 4,666 sad tweets, crawled via hashtags \cite{saravia-etal-2018-carer} to elicit responses and assess performance on unseen data.
To evaluate the empathetic levels expressed by these responses, we use a RoBERTa-based scoring model \cite{sharma-etal-2020-computational} that assigns a score between $0$ and $2$, inclusive, to each of the three ways of expressing empathy: emotional reaction (ER), interpretation (IP), and exploration (EX). We examine the mean ($\mu$) and standard deviation ($\sigma$) of these responses.

The summary statistics for empathy scores are shown in Table~\ref{tab:empathy score}. While IP and EX remain low for both models with and after fine-tuning, there is a notable 21\% increase in mean ER score accompanied by a 4\% decrease in the standard deviation. This indicates that fine-tuning on our corpus has enabled the Mistral 7B model to produce empathy more consistently in the form of appropriate emotional reaction. 
Although the fine-tuned model have lower means for IP and EX, these scores are already very low for the base model. Interestingly, similar trends were observed in CoE, as \citet{lee2023chain} reported a decrease in EX F1-score for all four types of CoE prompting (CBT, DBT, PCT, and RT), with an average drop of 12.53\%. 
This pattern suggests that a slight reduction in EX may be an inevitable trade-off for enhanced emotional reaction capabilities when employing our CoE-based corpus.
\begin{table}
  \centering
  \begin{tabular}{lcc}
    \hline
    Area & \textbf{Base} & \textbf{Fine-Tuned} \\
    \hline
    \textbf{ER}      & $\mu=1.16, \sigma=0.53$ & $\mu=1.40, \sigma=0.51$  \\
    \textbf{IP}      & $\mu=0.03, \sigma=0.24$ & $\mu=0.02, \sigma=0.20$  \\
    \textbf{EX}      & $\mu=0.11, \sigma=0.46$ & $\mu=0.04, \sigma=0.28$ \\\hline
  \end{tabular}
  \caption{Improvement in ER Empathy score of Mistral 7B after fine-tuning on SYNTHEMPATHY corpus. The empathy areas are emotional reaction (ER), interpretation (IP), and exploration (EX).}
  \label{tab:empathy score}
\end{table}
We have also manually inspected examples from the corpus and tracked the formation of each example across the entire multi-step process (examples in the Appendix Figures \ref{fig:cbt-all},\ref{fig:dbt-all},\ref{fig:pct-all},\ref{fig:rt-all}).

\section{Conclusion and Future Work}
We have created SYNTHEMPATHY, a novel, large-scale empathy corpus of 105k dialogues based on psychotherapy theories. We demonstrate that this corpus can enhance the emotional empathetic abilities of LLMs using an empathy scoring algorithm.
Beyond the corpus itself, we propose a step-by-step framework for constructing specialized corpora, as it can be generalized to downstream tasks beyond empathy.
The only components of our pipeline specifically tailored to empathy are the initial SAD dataset and CoE prompting. Given the availability of many domain-specific open datasets, CoE prompting can be substituted with domain-specific prompting methods as needed. For our future work, we plan to adapt our framework to low-resource areas, such as social norms.



\section*{Limitations}
Although our automatic corpus construction pipeline enables the creation of an entire empathy corpus internally, the trade-off involves replacing crowdsourcing with electricity consumption from a large amount of LLM inference. Most of the scenario brainstorming process was done by running inference with locally downloaded Llama 2 13B Chat on our machine with two NVIDIA L40 GPUs. However, our Llama 2 13B Chat inference did not use the full $300W$ TDP avaiable on L40. The other LLMs were smaller 7B models and required one GPU instead. During evaluation, we used one NVIDIA V100 GPU when fine-tuning Mistral 7B. The total GPU hours across all experiments spanned five days, with an average electricity consumption of $371.6W$ during the first three days and $152.4W$ for the remaining two days. This resulted in a total energy consumption of $34.1kWh$, which is around the average person's daily electricity usage in the United States ($29kWh$). Since it is a one-off cost for our pipeline, energy consumption does not pose a severe problem  and presents a more efficient alternative to eliminating the need for crowdsourcing.


\section*{Ethics Statement}
Although no unethical practices occurred during the construction of the SYNTHEMPATHY corpus, addressing its ethical implications is crucial given its connection to psychotherapy approaches and potential use in fine-tuning chatbots for individuals with mental health concerns.
Since the SYNTHEMPATHY corpus was built through an automated pipeline, there is a risk of inappropriate or sensitive topics entering the dataset via LLM output. 
To mitigate this risk, we scan the entire corpus to rigorously review and check for any presence from a dictionary of any sensitive words.
We removed 457 examples containing one or more of these sensitive words.

All supplementary datasets we used throughout the paper are open-sourced and publicly available. The inference code we adapted from Meta's Llama models are open source and our use aligns with their responsible use guide. The Unsloth\footnote{\url{https://unsloth.ai/}} code we adapt to fine-tune Mistral 7B is open sourced on their public GitHub\footnote{\url{https://github.com/unslothai/unsloth}} repository and they state that their notebooks can be used to fine-tune at no cost. SYNTHEMPATHY is an open-sourced corpus created to advance research in the empathy domain of LLMs ensuring full compliance with all terms of use.

\section*{Acknowledgments}
We thank the reviewers for their valuable comments and suggestions. Further details on acknowledgments will be provided upon completion of the review process to maintain the anonymity during the double-blind review.

\bibliography{anthology,custom}

\begin{thebibliography}{18}
\providecommand{\natexlab}[1]{#1}

\bibitem[{Chen et~al.(2024)Chen, Chen, Kulkarni, Lin, Pang, Tadimeti, Shin, and Hirschberg}]{chen2024detecting}
Run Chen, Haozhe Chen, Anushka Kulkarni, Eleanor Lin, Linda Pang, Divya Tadimeti, Jun Shin, and Julia Hirschberg. 2024.
\newblock Detecting empathy in speech.
\newblock In \emph{Proc. INTERSPEECH 2024}.

\bibitem[{Chen et~al.(2023)Chen, Xing, Lin, Zheng, Wang, Liu, and Xu}]{chen-etal-2023-soulchat}
Yirong Chen, Xiaofen Xing, Jingkai Lin, Huimin Zheng, Zhenyu Wang, Qi~Liu, and Xiangmin Xu. 2023.
\newblock \href {https://doi.org/10.18653/v1/2023.findings-emnlp.83} {{S}oul{C}hat: Improving {LLM}s{'} empathy, listening, and comfort abilities through fine-tuning with multi-turn empathy conversations}.
\newblock In \emph{Findings of the Association for Computational Linguistics: EMNLP 2023}, pages 1170--1183, Singapore. Association for Computational Linguistics.

\bibitem[{Fuller et~al.(2021)Fuller, Kamans, van Vuuren, Wolfensberger, and de~Jong}]{fuller2021conceptualizing}
Melissa Fuller, Elanor Kamans, Mark van Vuuren, Marca Wolfensberger, and Menno~DT de~Jong. 2021.
\newblock Conceptualizing empathy competence: a professional communication perspective.
\newblock \emph{Journal of business and technical communication}, 35(3):333--368.

\bibitem[{Girotra et~al.(2023)Girotra, Meincke, Terwiesch, and Ulrich}]{Girotra2023IdeasAD}
Karan Girotra, Lennart Meincke, Christian Terwiesch, and Karl~T. Ulrich. 2023.
\newblock \href {https://api.semanticscholar.org/CorpusID:260467886} {Ideas are dimes a dozen: Large language models for idea generation in innovation}.
\newblock \emph{SSRN Electronic Journal}.

\bibitem[{Goel et~al.(2021)Goel, Susan, Vashisht, and Dhanda}]{goel2021emotion}
Raman Goel, Seba Susan, Sachin Vashisht, and Armaan Dhanda. 2021.
\newblock Emotion-aware transformer encoder for empathetic dialogue generation.
\newblock In \emph{2021 9th International Conference on Affective Computing and Intelligent Interaction Workshops and Demos (ACIIW)}, pages 1--6. IEEE.

\bibitem[{Jani et~al.(2012)Jani, Blane, and Mercer}]{jani2012role}
Bhautesh~Dinesh Jani, David~N Blane, and Stewart~W Mercer. 2012.
\newblock The role of empathy in therapy and the physician-patient relationship.
\newblock \emph{Complementary Medicine Research}, 19(5):252--257.

\bibitem[{Lee et~al.(2022{\natexlab{a}})Lee, Lee, and Gan}]{lee2022improving}
Jing~Yang Lee, Kong~Aik Lee, and Woon~Seng Gan. 2022{\natexlab{a}}.
\newblock Improving contextual coherence in variational personalized and empathetic dialogue agents.
\newblock In \emph{ICASSP 2022-2022 IEEE International Conference on Acoustics, Speech and Signal Processing (ICASSP)}, pages 7052--7056. IEEE.

\bibitem[{Lee et~al.(2022{\natexlab{b}})Lee, Ippolito, Nystrom, Zhang, Eck, Callison-Burch, and Carlini}]{lee-etal-2022-deduplicating}
Katherine Lee, Daphne Ippolito, Andrew Nystrom, Chiyuan Zhang, Douglas Eck, Chris Callison-Burch, and Nicholas Carlini. 2022{\natexlab{b}}.
\newblock \href {https://doi.org/10.18653/v1/2022.acl-long.577} {Deduplicating training data makes language models better}.
\newblock In \emph{Proceedings of the 60th Annual Meeting of the Association for Computational Linguistics (Volume 1: Long Papers)}, pages 8424--8445, Dublin, Ireland. Association for Computational Linguistics.

\bibitem[{Lee et~al.(2023)Lee, Lee, Shin, Bae, and Hahn}]{lee2023chain}
Yoon~Kyung Lee, Inju Lee, Minjung Shin, Seoyeon Bae, and Sowon Hahn. 2023.
\newblock \href {https://arxiv.org/abs/2311.04915} {Chain of empathy: Enhancing empathetic response of large language models based on psychotherapy models}.
\newblock \emph{Preprint}, arXiv:2311.04915.

\bibitem[{Lucas et~al.(2018)Lucas, Boberg, Traum, Artstein, Gratch, Gainer, Johnson, Leuski, and Nakano}]{9473685}
Gale~M. Lucas, Jill Boberg, David Traum, Ron Artstein, Jonathan Gratch, Alesia Gainer, Emmanuel Johnson, Anton Leuski, and Mikio Nakano. 2018.
\newblock Getting to know each other: The role of social dialogue in recovery from errors in social robots.
\newblock In \emph{2018 13th ACM/IEEE International Conference on Human-Robot Interaction (HRI)}, pages 344--351.

\bibitem[{Mauriello et~al.(2021)Mauriello, Lincoln, Hon, Simon, Jurafsky, and Paredes}]{10.1145/3411763.3451799}
Matthew~Louis Mauriello, Thierry Lincoln, Grace Hon, Dorien Simon, Dan Jurafsky, and Pablo Paredes. 2021.
\newblock \href {https://doi.org/10.1145/3411763.3451799} {Sad: A stress annotated dataset for recognizing everyday stressors in sms-like conversational systems}.
\newblock In \emph{Extended Abstracts of the 2021 CHI Conference on Human Factors in Computing Systems}, CHI EA '21, New York, NY, USA. Association for Computing Machinery.

\bibitem[{Moss et~al.(2023)Moss, Rosenzweig, Robinson, Jaffe, and Litman}]{moss2023ethical}
Aaron~J Moss, Cheskie Rosenzweig, Jonathan Robinson, Shalom~N Jaffe, and Leib Litman. 2023.
\newblock Is it ethical to use mechanical turk for behavioral research? relevant data from a representative survey of mturk participants and wages.
\newblock \emph{Behavior Research Methods}, 55(8):4048--4067.

\bibitem[{Rashkin et~al.(2019)Rashkin, Smith, Li, and Boureau}]{rashkin-etal-2019-towards}
Hannah Rashkin, Eric~Michael Smith, Margaret Li, and Y-Lan Boureau. 2019.
\newblock \href {https://doi.org/10.18653/v1/P19-1534} {Towards empathetic open-domain conversation models: A new benchmark and dataset}.
\newblock In \emph{Proceedings of the 57th Annual Meeting of the Association for Computational Linguistics}, pages 5370--5381, Florence, Italy. Association for Computational Linguistics.

\bibitem[{Saravia et~al.(2018)Saravia, Liu, Huang, Wu, and Chen}]{saravia-etal-2018-carer}
Elvis Saravia, Hsien-Chi~Toby Liu, Yen-Hao Huang, Junlin Wu, and Yi-Shin Chen. 2018.
\newblock \href {https://doi.org/10.18653/v1/D18-1404} {{CARER}: Contextualized affect representations for emotion recognition}.
\newblock In \emph{Proceedings of the 2018 Conference on Empirical Methods in Natural Language Processing}, pages 3687--3697, Brussels, Belgium. Association for Computational Linguistics.

\bibitem[{Sharma et~al.(2020)Sharma, Miner, Atkins, and Althoff}]{sharma-etal-2020-computational}
Ashish Sharma, Adam Miner, David Atkins, and Tim Althoff. 2020.
\newblock \href {https://doi.org/10.18653/v1/2020.emnlp-main.425} {A computational approach to understanding empathy expressed in text-based mental health support}.
\newblock In \emph{Proceedings of the 2020 Conference on Empirical Methods in Natural Language Processing (EMNLP)}, pages 5263--5276, Online. Association for Computational Linguistics.

\bibitem[{Webb and Tangney(2022)}]{webb2022too}
Margaret~A Webb and June~P Tangney. 2022.
\newblock Too good to be true: Bots and bad data from mechanical turk.
\newblock \emph{Perspectives on Psychological Science}, page 17456916221120027.

\bibitem[{Wei et~al.(2022)Wei, Wang, Schuurmans, Bosma, Xia, Chi, Le, Zhou et~al.}]{wei2022chain}
Jason Wei, Xuezhi Wang, Dale Schuurmans, Maarten Bosma, Fei Xia, Ed~Chi, Quoc~V Le, Denny Zhou, et~al. 2022.
\newblock Chain-of-thought prompting elicits reasoning in large language models.
\newblock \emph{Advances in neural information processing systems}, 35:24824--24837.

\bibitem[{Welivita et~al.(2020)Welivita, Xie, and Pu}]{welivita2020finegrained}
Anuradha Welivita, Yubo Xie, and Pearl Pu. 2020.
\newblock \href {https://arxiv.org/abs/2012.13624} {Fine-grained emotion and intent learning in movie dialogues}.
\newblock \emph{Preprint}, arXiv:2012.13624.

\end{thebibliography}

\appendix

\section{Appendix}
\label{sec:appendix}

\subsection{Psychotherpy Theory Behind CoE Prompting}
CBT is a therapy style where the therapist tries to correct any misconceptions or catastrophic thoughts that the client has. Thus, CBT prompting appends to the original explanation that the client is overestimating the severity of the situation. This urges the LLM to respond by defusing the cognitive blind spot by empathetically suggesting alternative ways of thinking. DBT prompting appends to the original explanation that the client is having difficulties controlling their emotions. This in turn makes the LLM gear its response towards providing as much empathy as possible so that the client can become emotionally stable again. PCT prompting adds that the client is confused and unable to understand themselves to the base prompt. Finally, RT prompting adds that the client does not know where the root of their problems hides which makes the LLM focus on giving potential solutions while still being empathetic. In short, all four prompting methods are designed to maximize empathy by pinning down on a specific way in which therapists show empathy.

\subsection{LLM Hyperparameter Tuning}
\label{appendix:HyperparameterTuning}
\begin{figure*}
  \centering
  \includegraphics[scale=0.4] {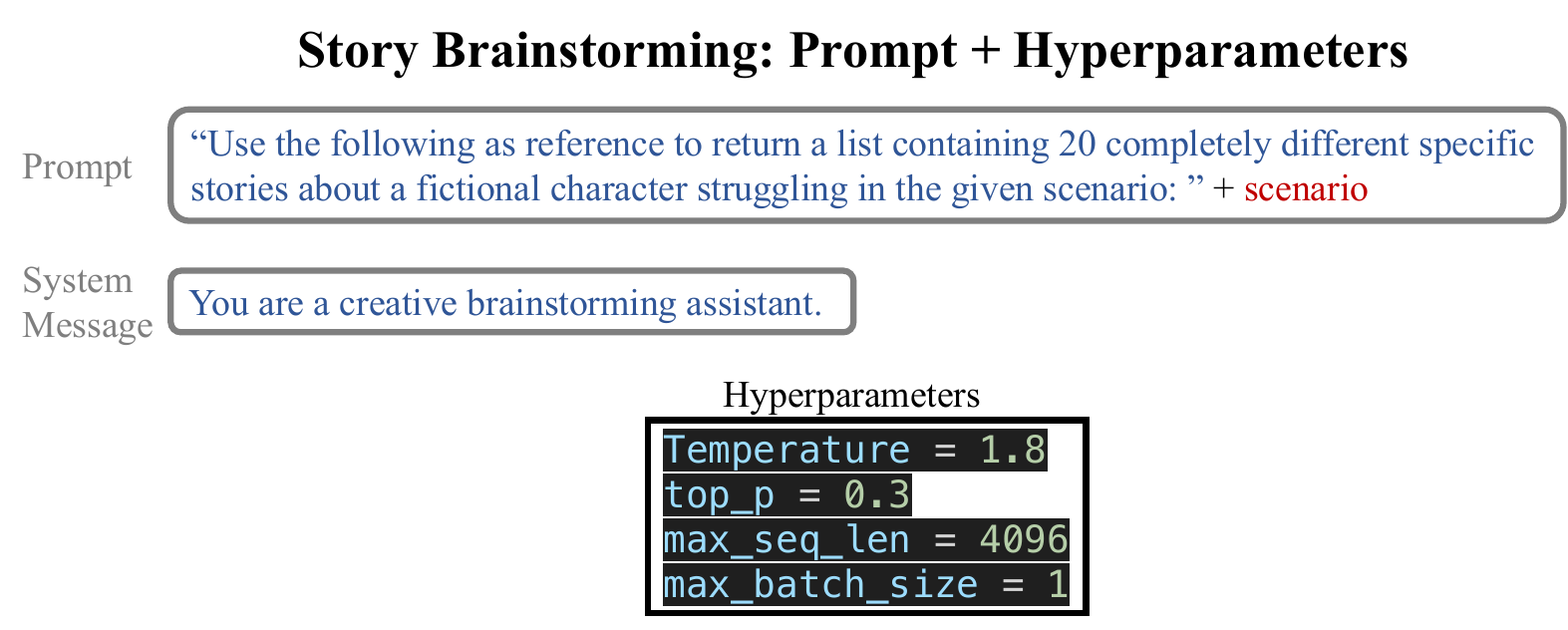}
  \caption{Prompts and Hyperparameters for First Step: Story Brainstorming}
  \label{fig:story-brainstorming-prompt}
\end{figure*}

\begin{figure*}
  \centering
  \includegraphics[scale=0.4] {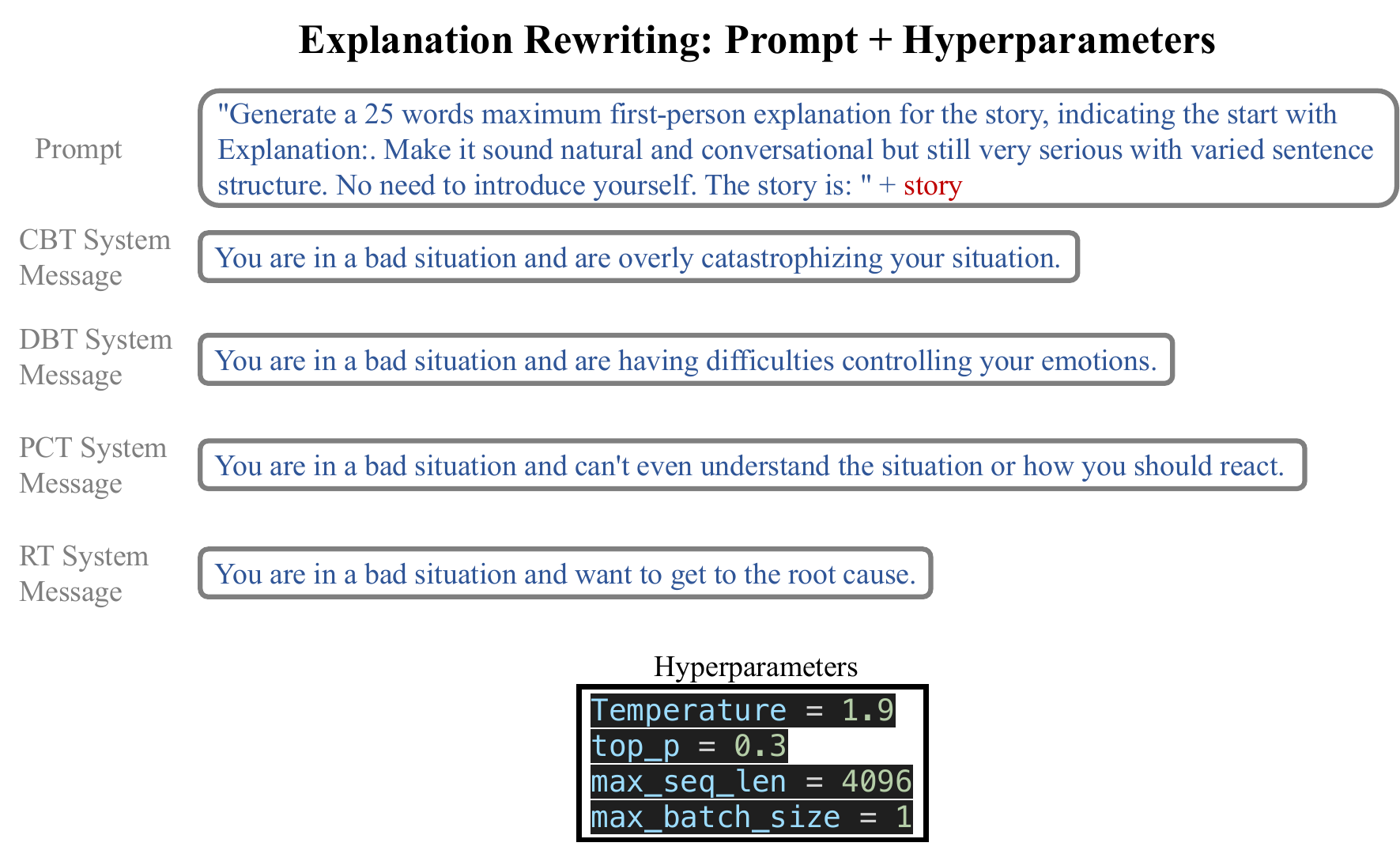}
  \caption{Prompts and Hyperparameters for Second Step: Explanation Rewriting}
  \label{fig:explanation-rewriting-prompt}
\end{figure*}

\begin{figure*}
  \centering
  \includegraphics[scale=0.4] {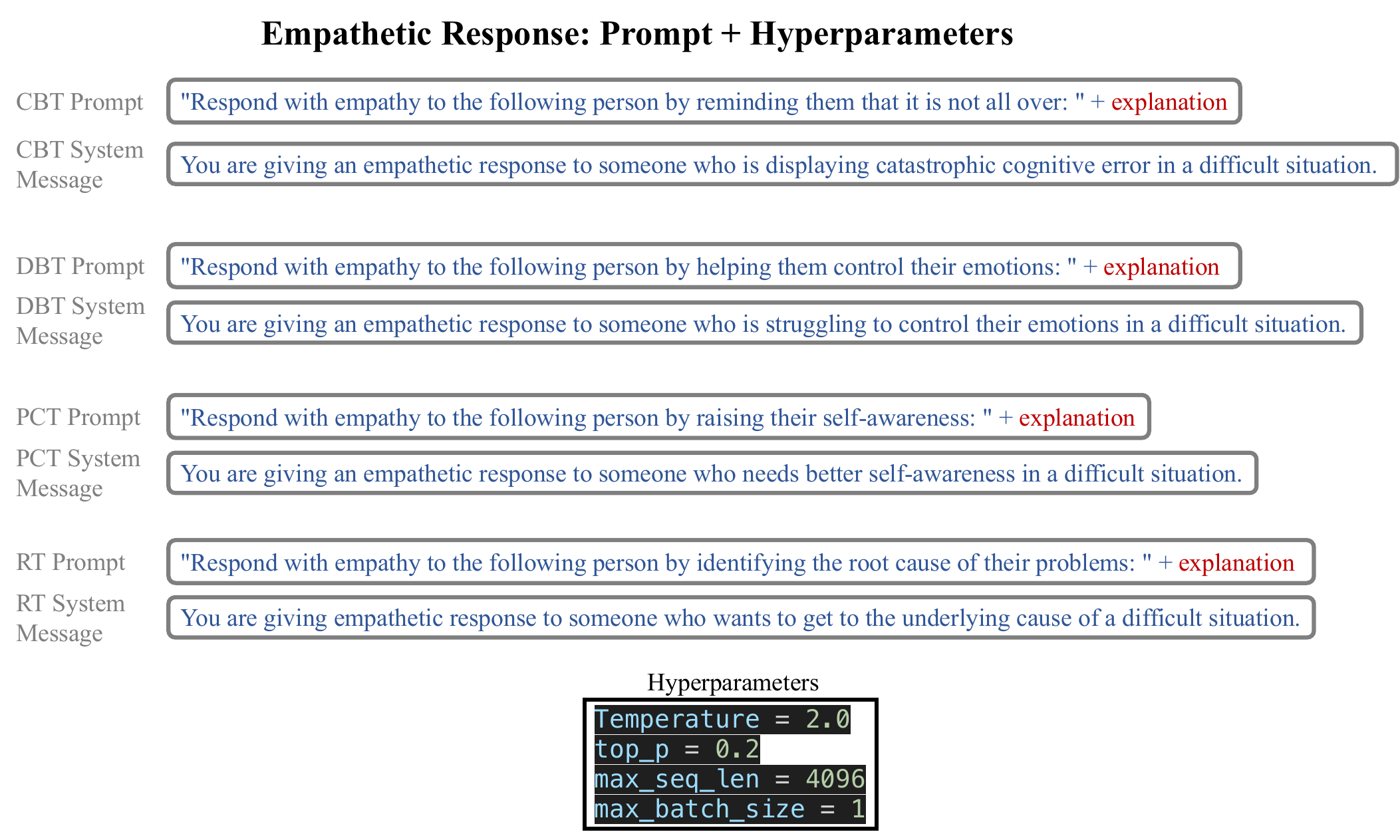}
  \caption{Prompts and Hyperparameters for Third Step: Empathetic Response}
  \label{fig:empathetic-response-prompt}
\end{figure*}

We use a grid search with step size $0.05$ to find the optimal values of \textit{temperature} and \textit{top\_p}. For each combination of the two hyperparameters we generate stories for $5$ scenarios and compare the outputs by inspection. For instance, for story brainstorming with Llama 2 13B Chat, we find that a relatively high \textit{temperature} of $1.8$ paired with a low $top\_p$ value of $0.3$ helps introduce as much randomness into the corpus as possible while keeping the LLM's responses coherent. We administer a similar grid search for explanation rewriting and empathetic response. Furthermore, we also do this for each of the other LLMs such as Gemma 7B and Mistral 7B.
Hyperparameters along with prompts are shown in Figures \ref{fig:story-brainstorming-prompt},\ref{fig:explanation-rewriting-prompt},\ref{fig:empathetic-response-prompt}.
\newpage 

\newpage

\begin{figure*}
  \centering
  \includegraphics[scale=0.5] {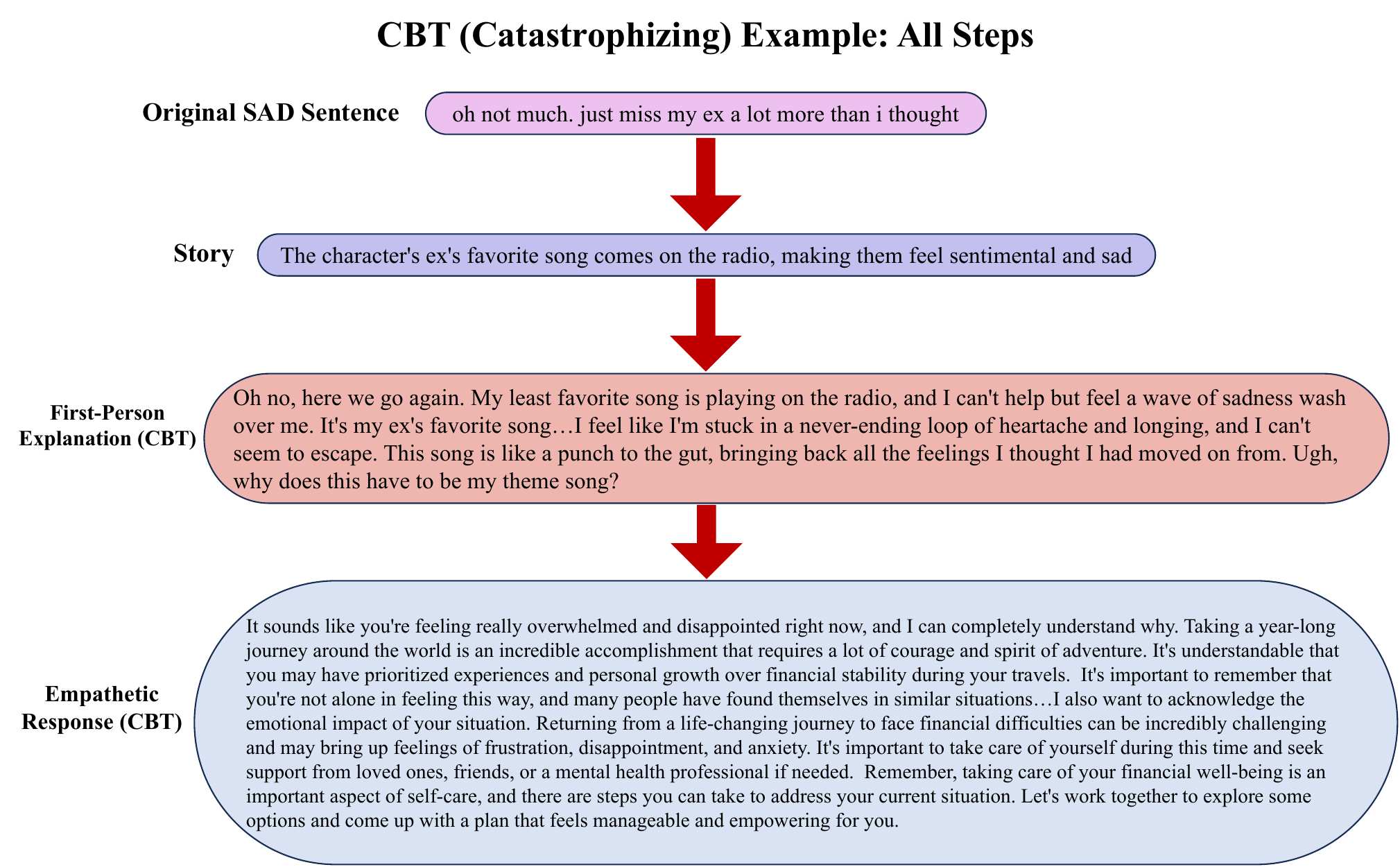}
  \caption{Random Sample for CBT}
  \label{fig:cbt-all}
\end{figure*}

\begin{figure*}
  \centering
  \includegraphics[scale=0.5] {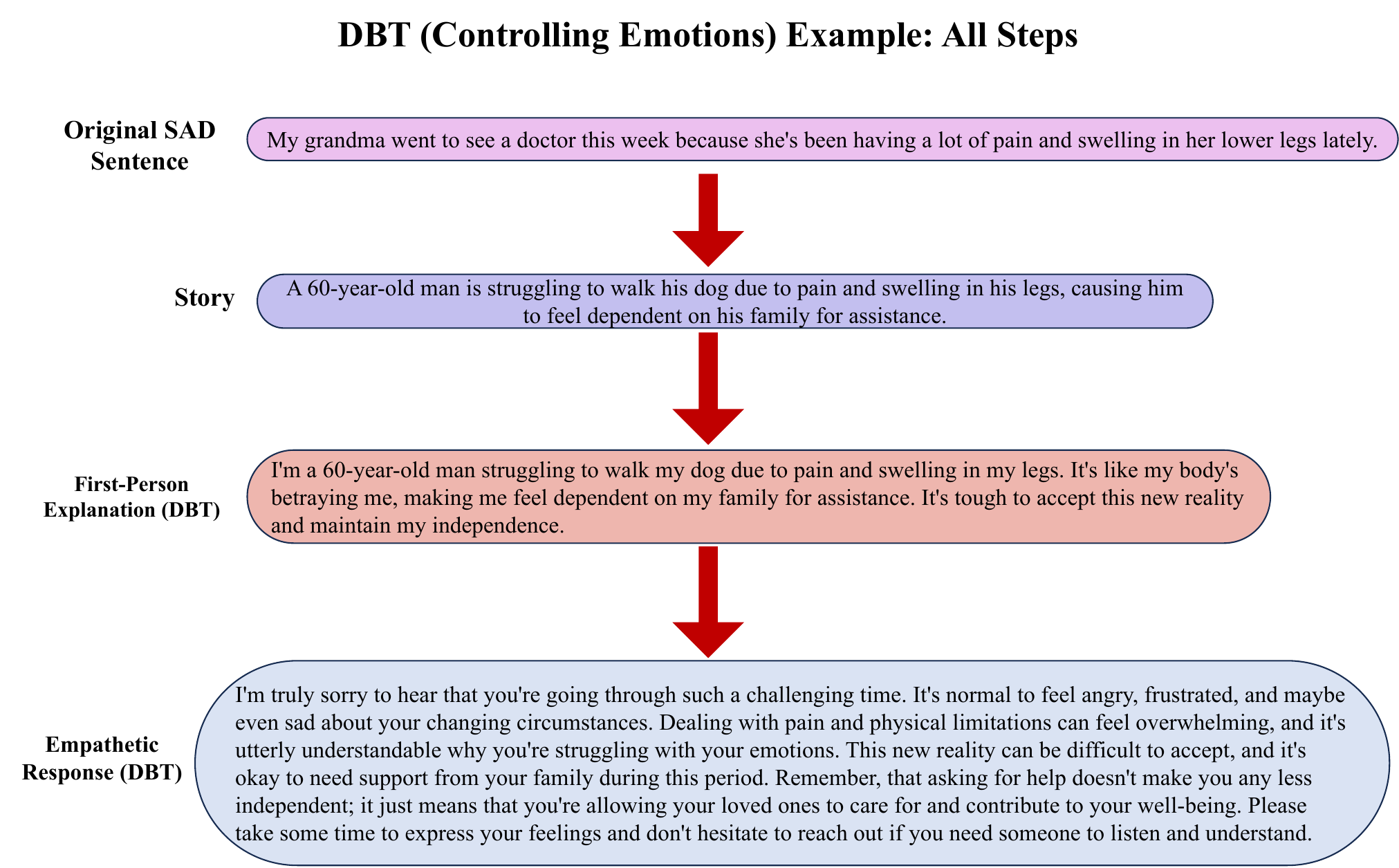}
  \caption{Random Sample for DBT}
  \label{fig:dbt-all}
\end{figure*}

\newpage

\begin{figure*}
  \centering
  \includegraphics[scale=0.5] {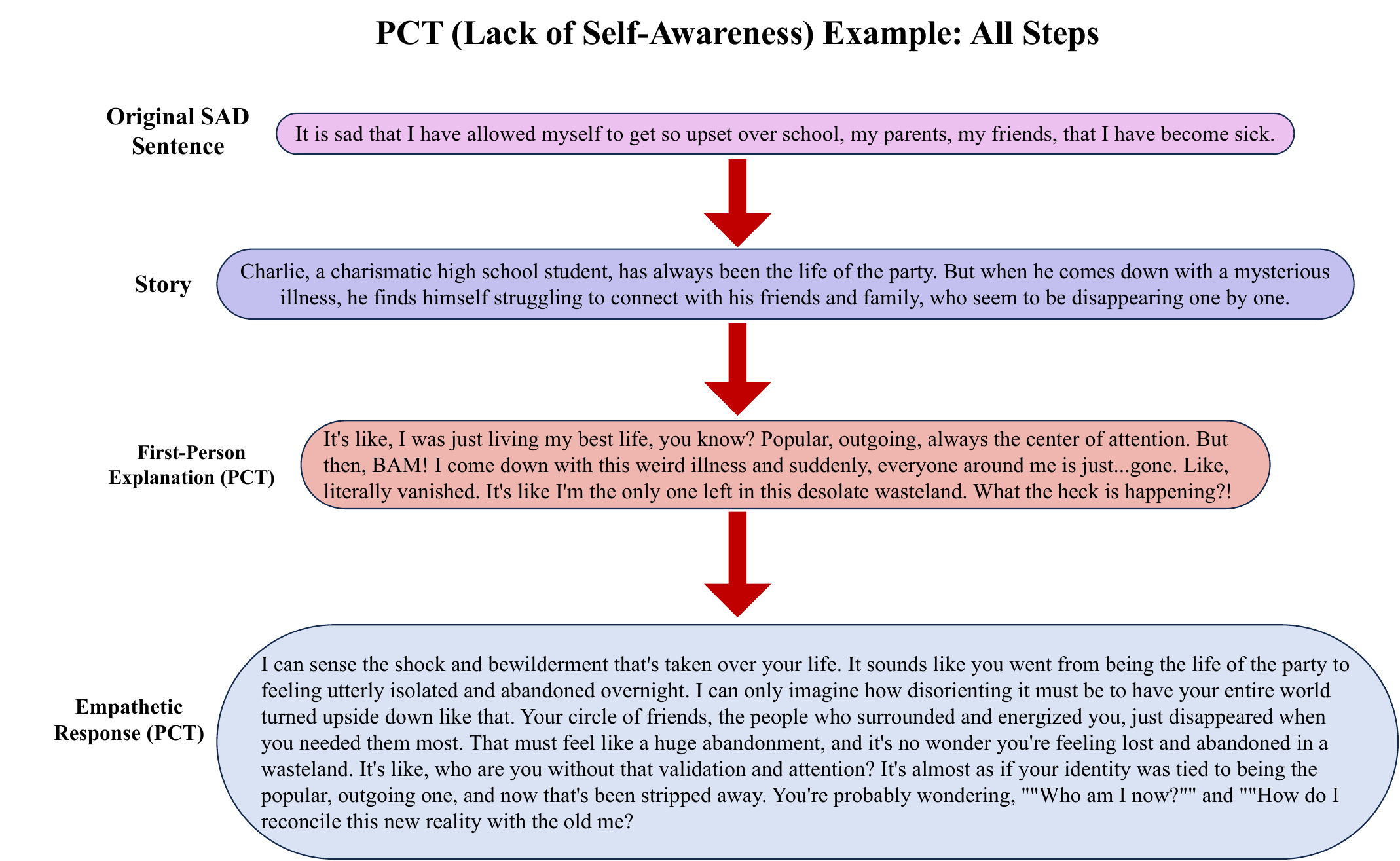}
  \caption{Random Sample for PCT}
  \label{fig:pct-all}
\end{figure*}

\begin{figure*}
  \centering
  \includegraphics[scale=0.5] {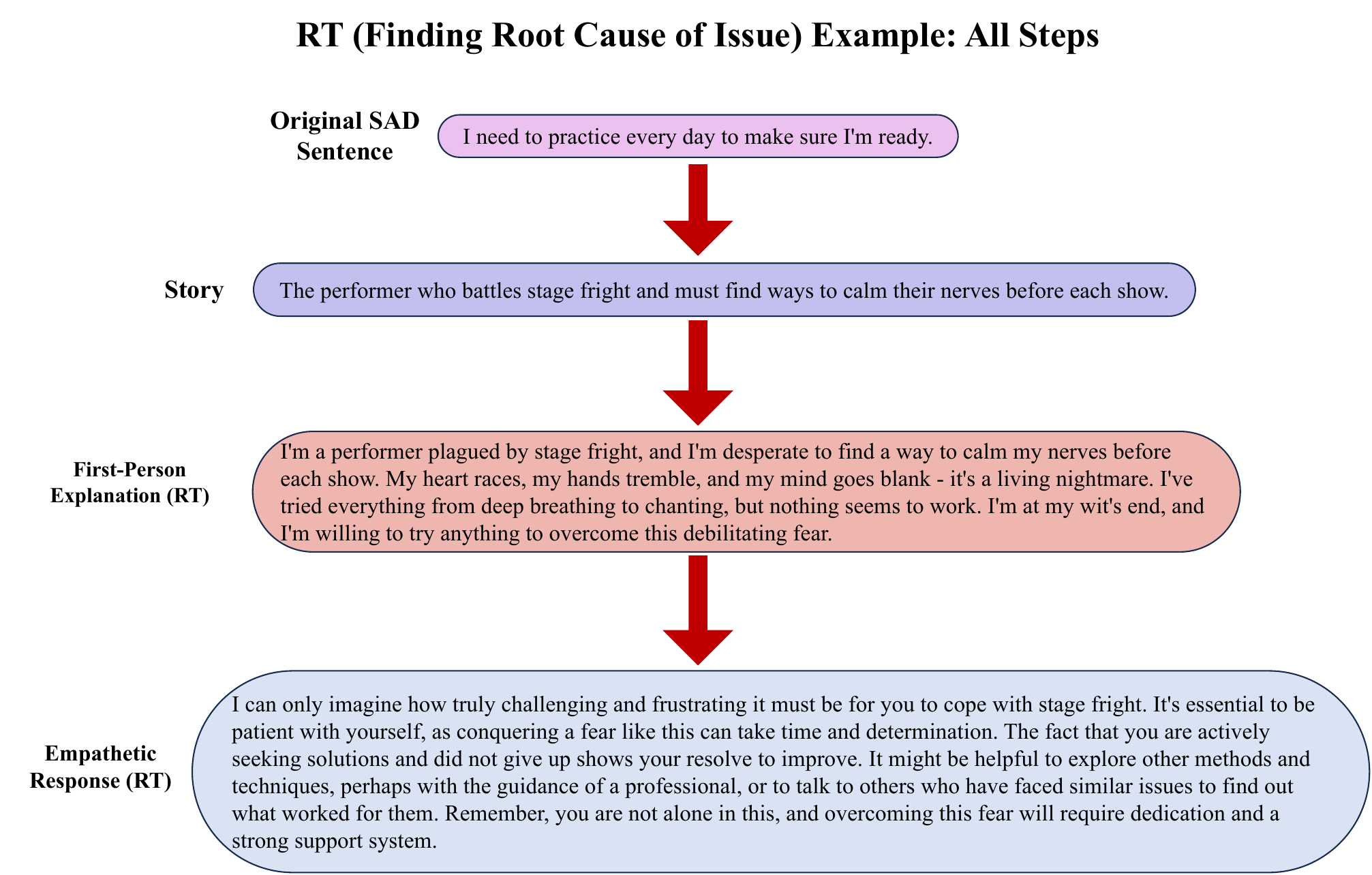}
  \caption{Random Sample for RT}
  \label{fig:rt-all}
\end{figure*}

\newpage

\begin{figure*}
  \centering
  \includegraphics[scale=0.8] {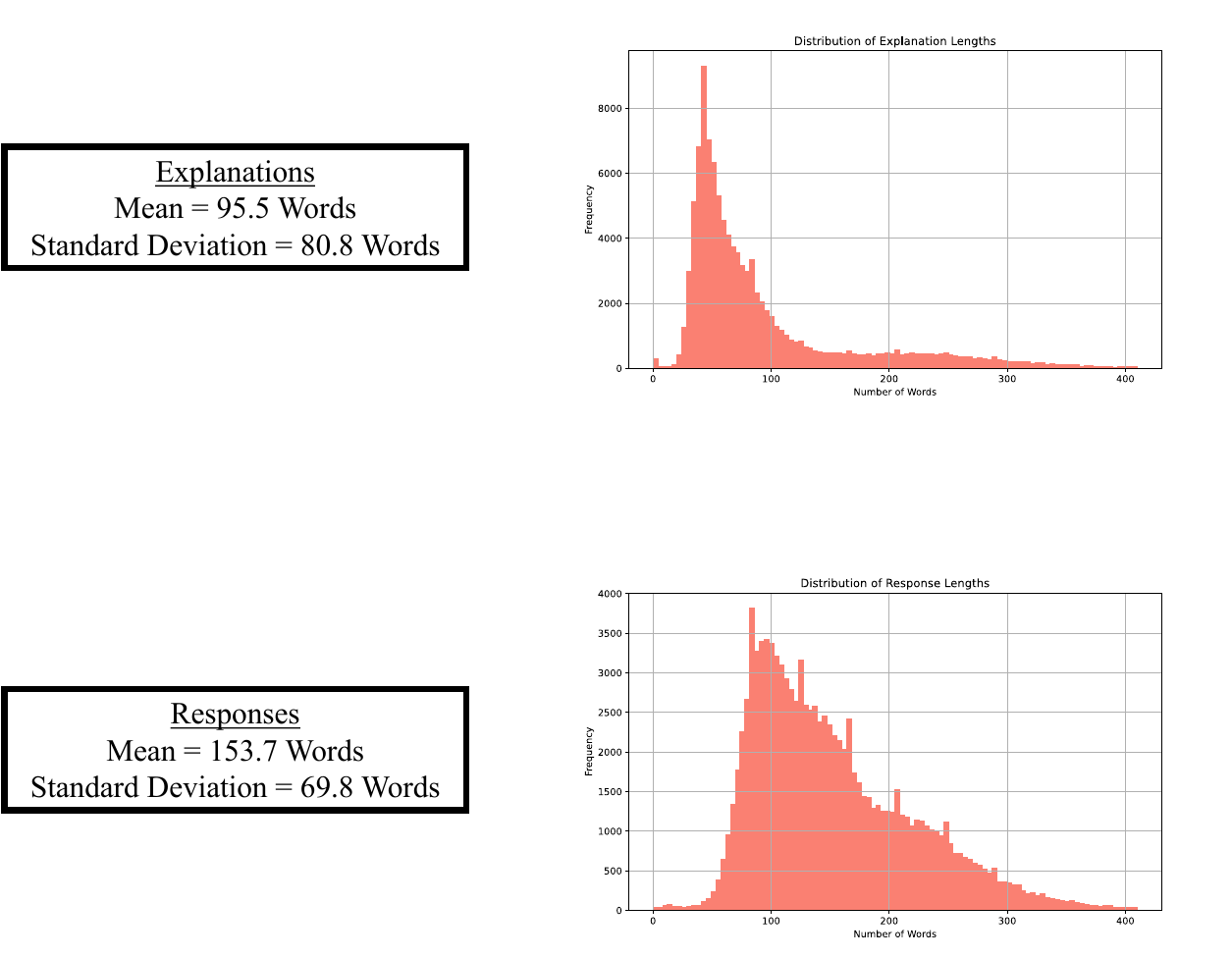}
  \caption{Distribution and Summary Statistics for Explanation and Responses in SYNTHEMPATHY}
  \label{fig:data-stats}
\end{figure*}

\end{document}